\documentclass[anonymous=false, %
               format=acmsmall, %
               nonacm=true]{acmart}

\usepackage[ruled]{algorithm2e} 
\usepackage{subcaption}
\newtheorem{assumption}{Assumption}
\usepackage{multirow}
\usepackage{soul}
\sethlcolor{yellow}

\begin{document}

\title{Gaussian Processes to speed up MCMC with automatic exploratory-exploitation effect}

\author{Alessio Benavoli}
\author{Jason Wyse}
\author{Arthur White}
\affiliation{%
  \institution{School of Computer Science and Statistics, Trinity College Dublin, Ireland}
}
\email{alessio.benavoli@tcd.ie}
\email{wyseja@tcd.ie}
\email{arwhite@tcd.ie}

\maketitle

\section{Introduction}
In this paper, we consider the problem of sampling from a posterior
$$
\pi({\boldsymbol \theta}|D)\propto p(D|{\boldsymbol \theta})p({\boldsymbol \theta}),
$$
where $D$ denotes data and ${\boldsymbol \theta} \in \Theta$ is a vector of unknown parameters, in the case  where the likelihood $p(D|{\boldsymbol \theta})$ is costly to evaluate. We discuss two-stage algorithms. In the first of these, we examine an adaptive Metropolis-Hastings (MH) algorithm~\cite{Hastings70, Robert2015} which employs an adaptively tuned Gaussian Process (GP) surrogate model at the first stage to filter out poor proposals. If a proposal is not filtered out, at the second stage a full (expensive) log-likelihood evaluation is carried out and used to decide whether it is accepted as the next state. Introduction of the first stage, constructed in this way, saves computation on poor proposals. A key contribution of this work is in the form of the acceptance probability in the first stage obtained by marginalising out the GP function. This makes the acceptance ratio  dependent  on the variance of the GP, which naturally results in an exploration-exploitation trade-off similar to the one of Bayesian Optimisation
\citep{BCD10}, which allows us to sample while learning the GP. We demonstrate that using this expectation serves as a useful filtering scheme. The second algorithm is a two-stage form of Metropolis adjusted Langevin algorithm (MALA)~\cite{Neal2011}. Here, we use GP as a surrogate for the log-likelihood function again, but in this case the GP is also used to approximate the gradient required for MALA updating, using a well known result that the gradient of a GP is also a GP~\cite{Rasmussen02}. Marginalizing out of the GP can also be performed in this instance.  

The approximation we use is 
\begin{equation}
\label{eq:GPapprox}
 \text{LL}(D|{\boldsymbol \theta}):= \ln p(D|{\boldsymbol \theta}) \approx  \widetilde{\text{LL}}_t(D|{\boldsymbol \theta}) \sim \text{GP}(\mu({\boldsymbol \theta}|\mathcal{I}_t),k({\boldsymbol \theta},{\boldsymbol \theta}^*|\mathcal{I}_t))
\end{equation}
where $\mathcal{I}_t$ denotes the set of $t$ full evaluations of the log-likelihood by the current iteration, and $\boldsymbol \theta^*$ collectively denotes the parameter values at which these evaluations were made.
Adaptive tuning of the GP surrogate is accomplished through use of the collection $\mathcal{I}_t$ of full evaluations of the log-likelihood. We argue that the tuning schedule we suggest satisfies diminishing adaptation~\cite{roberts_rosenthal_2007} and hence will ensure correct sampling from the true target $\pi(\boldsymbol\theta|D)$.

Within the Markov chain Monte Carlo (MCMC) literature, there has been much interest in recent years, in the use of proxy quantities for the target measure evaluations from different aspects. Approaches using noisy approximations to an invariant transition kernel~\cite{Andrieu2009, Alquier16} have gained much interest. The work here assumes that the log-likelihood, though maybe expensive, can be computed, and is thus more aligned to the work of~\citet{rasmussen2003gaussian,Christen2005,Sherlock2017,li2019neural,fielding2011efficient,bliznyuk2012local,joseph2012bayesian}, involving ideas from delayed acceptance MCMC. The key difference is that we do not carry out pre-computation of the GP prior to running the algorithm, investigating adaptation of the GP on the fly using key results from the adaptive MCMC literature~\cite{roberts_rosenthal_2007} to ensure convergence to the true target. 

We present the two stage MH algorithm in Section~\ref{sec:two-stage-description}. Section~\ref{sec:MALA} follows by introducing a the two-stage MALA algorithm building on these ideas. Section~\ref{sec:examples} explores a range of examples and demonstrates the merits of the filtering step, with a discussion of potential drawbacks.

\section{Two Stage Adaptive Metropolis-Hastings via GP approximation} \label{sec:two-stage-description}

We combine the MH algorithm with a GP model which approximates the log-likelihood. In cases where the log-likelihood is expensive to compute, the GP model can be used in a pre-filtering step to determine proposals for which a full computation of the log-likelihood might well lead to an acceptance~\cite{Christen2005}.  %
At each iteration of the algorithm, the
first stage, uses a GP to deliver an approximate log-likelihood evaluation. The GP is based on a collection $\mathcal{I}_t$ of previous full evaluations of the log-likelihood. A propsal is made from the current state and then the usual MH acceptance probability is computed using the approximated  log-likelihood (this step is computationally inexpensive). If the proposal is accepted in this first stage, then it goes to the second stage, where another acceptance probability is computed, but this time, based on the full costly evaluation of the log-likelihood. The resulting evaluation of the log-likelihood is then appended to $\mathcal{I}_t$, resulting in $\mathcal{I}_{t+1}$.

Before giving a full description of the algorithm, we introduce some notation and give an explicit definition of $\mathcal{I}_t$: 
\begin{itemize}
 \item $\mathcal{S}_k$ denotes the points sampled up to the iteration $k$ of the algorithm;
 \item ${\boldsymbol \theta}^{(k)}$  denotes the most recent element in $\mathcal{S}_k$ and ${\boldsymbol \theta}^*$ denotes the proposed state
 \item $\mathcal{I}_t=\{({\boldsymbol \theta}^{(i)},\text{LL}(D|{\boldsymbol \theta}^{(i)})): ~~i=1,\dots,t\}$ denotes the $t\leq k$ exact likelihood evaluations performed  up to iteration $k$.
\end{itemize}
 We use a noise free GP as a surrogate model for the log-likelihood and denote by $\text{GP}_k(\mu({\boldsymbol \theta}|\mathcal{I}_t),k({\boldsymbol \theta},{\boldsymbol \theta}|\mathcal{I}_t))$ the posterior GP at the iteration $k$ conditioned on the collection $\mathcal{I}_t$.
We use $\widetilde{\text{LL}}_k({\boldsymbol \theta})$
 to denote the GP-distributed log-likelihood. 
 We choose the parameters of the GP to satisfy the following exact interpolation property. 
 \begin{assumption}
 \label{as1}
 The  prior mean function and prior covariance function of the GP are selected to guarantee exact interpolation:  %
 $$
 \mu({\boldsymbol \theta}^{(i)}|\mathcal{I}_t)=\text{LL}(D|{\boldsymbol \theta}^{(i)}),~~~~~~~\qquad \qquad~~~~~~~~~~ k({\boldsymbol \theta}^{(i)},{\boldsymbol \theta}|\mathcal{I}_t)=0,
 $$
 for all ${\boldsymbol \theta}^{(i)}$ with a corresponding entry in $\mathcal{I}_t$ and $ {\boldsymbol \theta} \in \Theta$.\footnote{Any universal covariance function satisfies this property, for instance Squared Exponential.}
 \end{assumption}
 This means the predictions of the GP at the points ${\boldsymbol \theta}^{(i)} \in \mathcal{I}_t$ are exact and certain (zero (co)variance), which is a desirable property in a noise free regression problem.\footnote{This also guarantees consistency between the two stages: the denominator of \eqref{eq:alphastar}
 and \eqref{eq:alpha} is the same}.

The two stages of the MH algorithm are as follows.
\paragraph{Stage 1} Use the  predictive posterior GP (conditioned on the collection $\mathcal{I}_t$) to approximate the log-likelihood.
Define the first stage  acceptance probability: 
\begin{align}
\label{eq:alphastar}
\tilde{\alpha}^{(1)}({\boldsymbol \theta}^{(k)},{\boldsymbol \theta}^*)= 1 \wedge \frac{\exp( \widetilde{\text{LL}}_t({\boldsymbol \theta}^*))p({\boldsymbol \theta}^*)q({\boldsymbol \theta}^{(k)}|{\boldsymbol \theta}^*)}{\exp( \widetilde{\text{LL}}_t({\boldsymbol \theta}^{(k)}))p({\boldsymbol \theta}^{(k)})q({\boldsymbol \theta}^*|{\boldsymbol \theta}^{(k)})}
\end{align}
where $\widetilde{\text{LL}}_t(\cdot) \sim \text{GP}_k(\mu({\boldsymbol \theta}|\mathcal{I}_t),k({\boldsymbol \theta},{\boldsymbol \theta}|\mathcal{I}_t))$ and  we use the shorthand notation $a \wedge b = \min(a, b)$. Note that, because of   the exact interpolation property in Assumption \ref{as1}, it results that   $\widetilde{\text{LL}}_t({\boldsymbol \theta}^{(k)})=\text{LL}_t({\boldsymbol \theta}^{(k)})$.

The acceptance probability $\tilde{\alpha}^{(1)}({\boldsymbol \theta}^{(k)} ,{\boldsymbol \theta}^*)$ (respectively, $\tilde{\alpha}^{(1)}({{\boldsymbol \theta}^*,\boldsymbol \theta}^{(k)})$) depends on $\widetilde{\text{LL}}_t({\boldsymbol \theta}^*)-\widetilde{\text{LL}}_t({\boldsymbol \theta}^{(k)})$ (respectively, $-\widetilde{\text{LL}}_t({\boldsymbol \theta}^*)+\widetilde{\text{LL}}_t({\boldsymbol \theta}^{(k)})$ ) which is GP distributed. A key part of our approach involves marginalizing this dependence out by exploiting the following result.

\begin{proposition}
\label{prop:1}
 The distribution of $ e^{\widetilde{\text{LL}}_t({\boldsymbol \theta})}$ is
 $ \textit{Lognormal}\left(\mu({\boldsymbol \theta}|\mathcal{I}_t),k({\boldsymbol \theta},{\boldsymbol \theta}|\mathcal{I}_t)\right),
 $
and its mean is 
 \begin{equation}
\label{eq:meanlognorm}
 e^{\mu({\boldsymbol \theta}|\mathcal{I}_t) +\frac{1}{2} k({\boldsymbol \theta},{\boldsymbol \theta}|\mathcal{I}_t)}.
 \end{equation}
\end{proposition}
The proofs of this and the other Propositions are given in Appendix~\ref{app:proofs}.
By Assumption \ref{as1}, we have that $k({\boldsymbol \theta}^{(k)},{\boldsymbol \theta}^{(k)}|\mathcal{I}_t)=k({\boldsymbol \theta}^*,{\boldsymbol \theta}^{(k)}|\mathcal{I}_t)=0$ and, therefore, $\widetilde{\text{LL}}_t({\boldsymbol \theta}^*)$ and $\widetilde{\text{LL}}_t({\boldsymbol \theta}^{(k)})$ are sampled independently.
By exploiting Proposition \ref{prop:1}, we remove the dependence of the acceptance probability on  $\widetilde{\text{LL}}$ in \eqref{eq:alphastar}  resulting in the acceptance probability:
\begin{equation}
\label{eq:alphastar1}
\alpha^{(1)}({\boldsymbol \theta}^{(k)},{\boldsymbol \theta}^*)= 1 \wedge \frac{e^{\mu({\boldsymbol \theta}^*|\mathcal{I}_t) +\frac{1}{2}k({\boldsymbol \theta}^*,{\boldsymbol \theta}^*|\mathcal{I}_t)}p({\boldsymbol \theta}^*)q({\boldsymbol \theta}^{(k)} |{\boldsymbol \theta}^*)}{e^{\mu({\boldsymbol \theta}^{(k)}|\mathcal{I}_t)}p({\boldsymbol \theta^{(k)}})q({\boldsymbol \theta}^*|{\boldsymbol \theta}^{(k)})},
\end{equation}
where $\mu({\boldsymbol \theta}^{(k)}|\mathcal{I}_t)=
{\text{LL}}({\boldsymbol \theta}^{(k)})$ is the exact log-likelihood (by Assumption \ref{as1}).

 It can be seen that  
$\mu({\boldsymbol \theta}^*|\mathcal{I}_t) +\frac{1}{2}k({\boldsymbol \theta}^*,{\boldsymbol \theta}^*|\mathcal{I}_t)$ depends on the GP variance and, therefore, the acceptance probability is larger in regions where the GP uncertainty is large. Similar to the acquisition functions in Bayesian optimisation, this naturally results in an exploration-exploitation trade-off. However, our goal here is different, we aim to sample from the target distribution.

Therefore, given \eqref{eq:alphastar1}, in Stage 1, we accept ${\boldsymbol \theta}^*$ with probability $\alpha^{(1)}({\boldsymbol \theta}^{(k)},{\boldsymbol \theta}^*)$, otherwise ${\boldsymbol \theta}^{(k+1)}={\boldsymbol \theta}^{(k)}$. This defines the following transition kernel at Stage 1:
\begin{align}
\label{eq:starproposal}
Q^*_k(A|{\boldsymbol \theta}^{(k)})&=\int_{A} \alpha^{(1)}({\boldsymbol \theta}^{(k)},{\boldsymbol \theta}^*)q({\boldsymbol \theta}^*|{\boldsymbol \theta}^{(k)})d{\boldsymbol \theta}^*+I_A({\boldsymbol \theta})\int_{\Theta} (1-\alpha^{(1)}({\boldsymbol \theta}^{(k)},{\boldsymbol \theta}^*))d{\boldsymbol \theta}^*.
\end{align}
One can show that the above transition kernel satisfies the detailed balance property for the approximated target distribution $e^{\mu({\boldsymbol \theta}|\mathcal{I}_t) +\frac{1}{2}k({\boldsymbol \theta},{\boldsymbol \theta}|\mathcal{I}_t)}p({\boldsymbol \theta})$. 
\begin{proposition}
\label{prop:2}
The transition kernel \eqref{eq:starproposal} satisfies detailed balance.
\end{proposition}

We are not interested in the approximated target distribution, this is the reason we perform the second stage.

\paragraph{Stage 2.} At Stage 2, we perform another MH acceptance step, evaluating the exact  log-likelihood. Let ${\boldsymbol \theta}^*$ denote a  point  sampled from  $q^*_k({\boldsymbol \theta}^*|{\boldsymbol \theta}^{(k)}):=Q^*_k( d{\boldsymbol \theta}^*|{\boldsymbol \theta}^{(k)})$. Note that, ${\boldsymbol \theta}^*$ is either equal to the point ${\boldsymbol \theta}^*$ sampled at Stage 1 or to ${\boldsymbol \theta}^{(k)}$ if  ${\boldsymbol \theta}^*$  was rejected at Stage 1.

So, with probability
\begin{align}
\nonumber
\alpha^{(2)}({\boldsymbol \theta}^{(k)},{\boldsymbol \theta}^*)&= 1 \wedge \frac{\exp( \text{LL}(D|{\boldsymbol \theta}^*))p({\boldsymbol \theta}^*)q^*_k({\boldsymbol \theta}^{(k)}|{\boldsymbol \theta}^*)}{\exp( \text{LL}(D|{\boldsymbol \theta}^{(k)}))p({\boldsymbol \theta}^{(k)})q^*_k({{\boldsymbol \theta}^*|\boldsymbol \theta}^{(k)})}\\
\label{eq:alpha}
&= 1 \wedge \frac{\exp( \text{LL}(D|{\boldsymbol \theta}^*))p({\boldsymbol \theta}^*)q({\boldsymbol \theta}^{(k)}|{\boldsymbol \theta}^*)\alpha^{(1)}({{\boldsymbol \theta}^*,\boldsymbol \theta}^{(k)})}{\exp( \text{LL}(D|{\boldsymbol \theta}^{(k)}))p({\boldsymbol \theta}^{(k)})q({{\boldsymbol \theta}^*|\boldsymbol \theta}^{(k)})\alpha^{(1)}({\boldsymbol \theta}^{(k)},{\boldsymbol \theta}^*)},
\end{align}
we accept ${\boldsymbol \theta}^*$, otherwise ${\boldsymbol \theta}^{(k+1)}={\boldsymbol \theta}^{(k)}$.

The definition of $q^*_k$ means a rejection at Stage 1 always leads to a rejection at Stage 2, we do not  need to compute \eqref{eq:alpha} when \eqref{eq:alphastar} has led to a rejection.
When the sample ${\boldsymbol \theta}^*$ is accepted a Stage 1, we compute the full log-likelihood, update the set $\mathcal{I}_t$, and evaluate \eqref{eq:alpha}.

Overall the acceptance probability for a new point ${\boldsymbol \theta}^*$ is $\alpha^{(1)}({\boldsymbol \theta}^{(k)},{\boldsymbol \theta}^*)\alpha^{(2)}({\boldsymbol \theta}^{(k)},{\boldsymbol \theta}^*)$.
The overall two-stage algorithm preserves detailed balance with respect to the posterior distribution and this follows directly by \eqref{eq:alpha}, which is a standard MH acceptance step with proposal $q_k({\boldsymbol \theta}^{(k)}|{\boldsymbol \theta}^*)\alpha^{(1)}({{\boldsymbol \theta}^*,\boldsymbol \theta}^{(k)})$.

\paragraph{Convergence analysis} To prove the convergence to the target distribution, it is enough to show that the overall (two-stage) transition kernel  $P_{t}(\cdot|{\boldsymbol \theta})$ 
satisfies the \textit{Diminishing Adaptation} condition in~\citet{roberts_rosenthal_2007}:
\begin{equation}
 \label{eq:diminish}
 \lim\limits_{t \rightarrow \infty} \sup_{{\boldsymbol \theta} \in \Theta} ||P_{t}(\cdot|{\boldsymbol \theta}) -P_{t-1}(\cdot|{\boldsymbol \theta})||=0 ~~\text{  in probability}.
\end{equation}
and the \textit{Bounded Convergence condition} which is generally
satisfied under some regularity conditions of $\Theta$ and the target distribution.  
The adaptivity  in our two-stage algorithm is due to the GP and diminishing adaptation follows by this property of the posterior predictive variance. 

\begin{proposition}
 \label{prop:3}
 For fixed hypeprameters, the surrogated model satisfies
 this property:
 $k({\boldsymbol \theta}^*,{\boldsymbol \theta}^*|\mathcal{I}_t)< k({\boldsymbol \theta}^*,{\boldsymbol \theta}^*|\mathcal{I}_{t-1})$.
\end{proposition}

\begin{figure}
\begin{subfigure}{.32\textwidth}
  \centering
  \includegraphics[width=.8\linewidth]{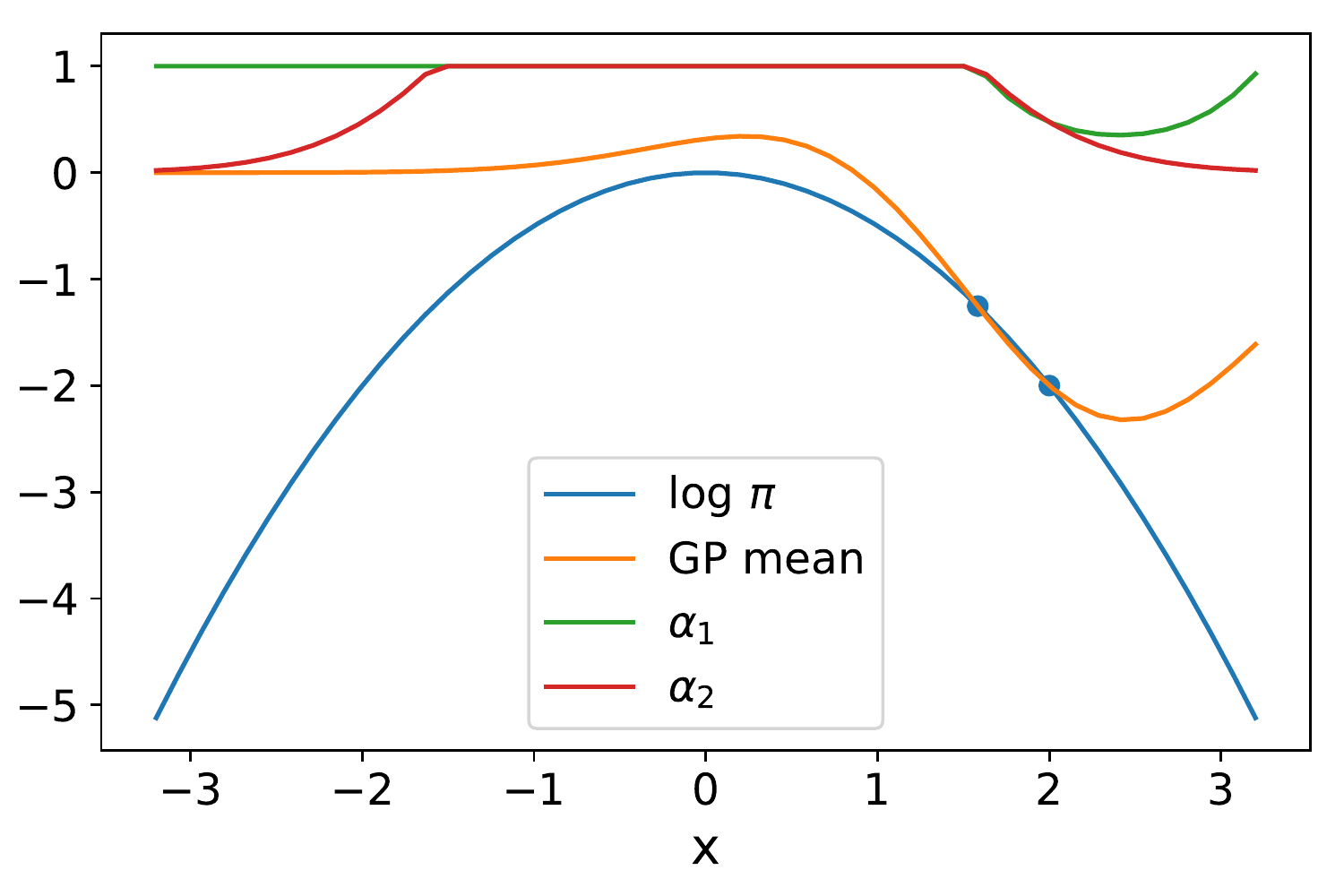}
  \caption{}
  \label{fig:sfig1}
\end{subfigure}%
\begin{subfigure}{.32\textwidth} 
  \centering
  \includegraphics[width=.8\linewidth]{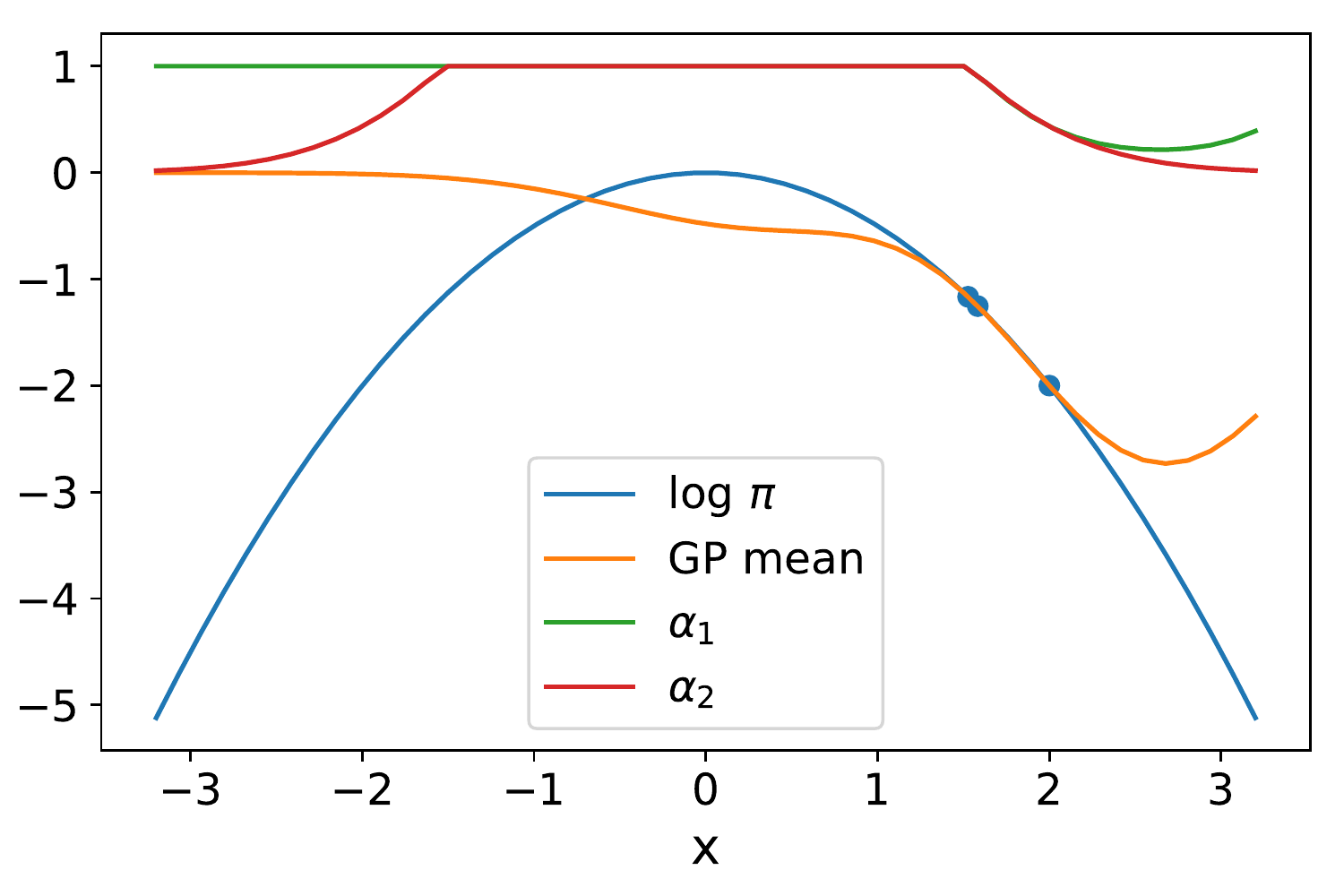}
  \caption{}
  \label{fig:sfig2} 
\end{subfigure}
\begin{subfigure}{.32\textwidth}
  \centering
  \includegraphics[width=.8\linewidth]{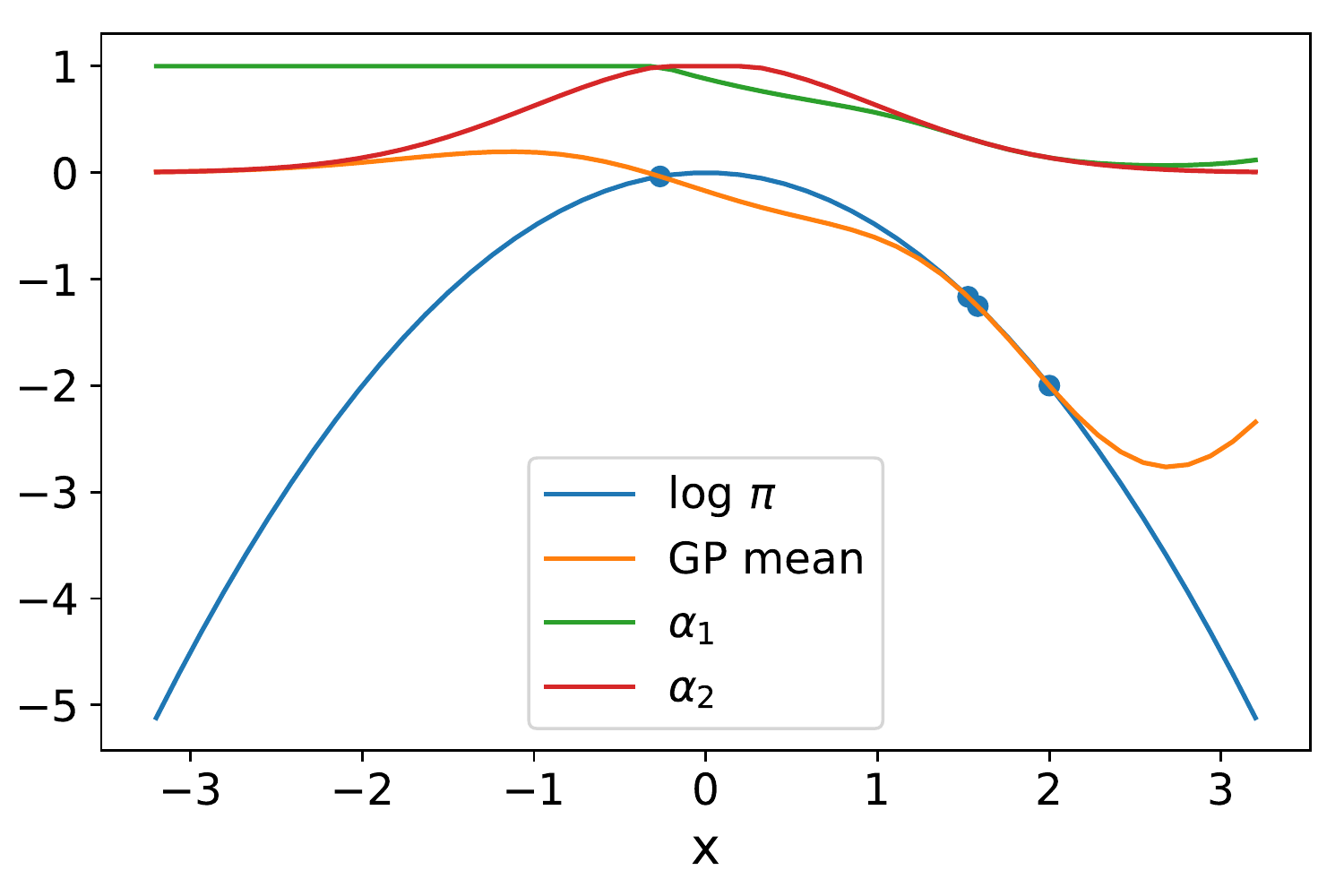}
  \caption{}
  \label{fig:sfig3}
\end{subfigure}\\
\begin{subfigure}{.32\textwidth}    
  \centering
  \includegraphics[width=.8\linewidth]{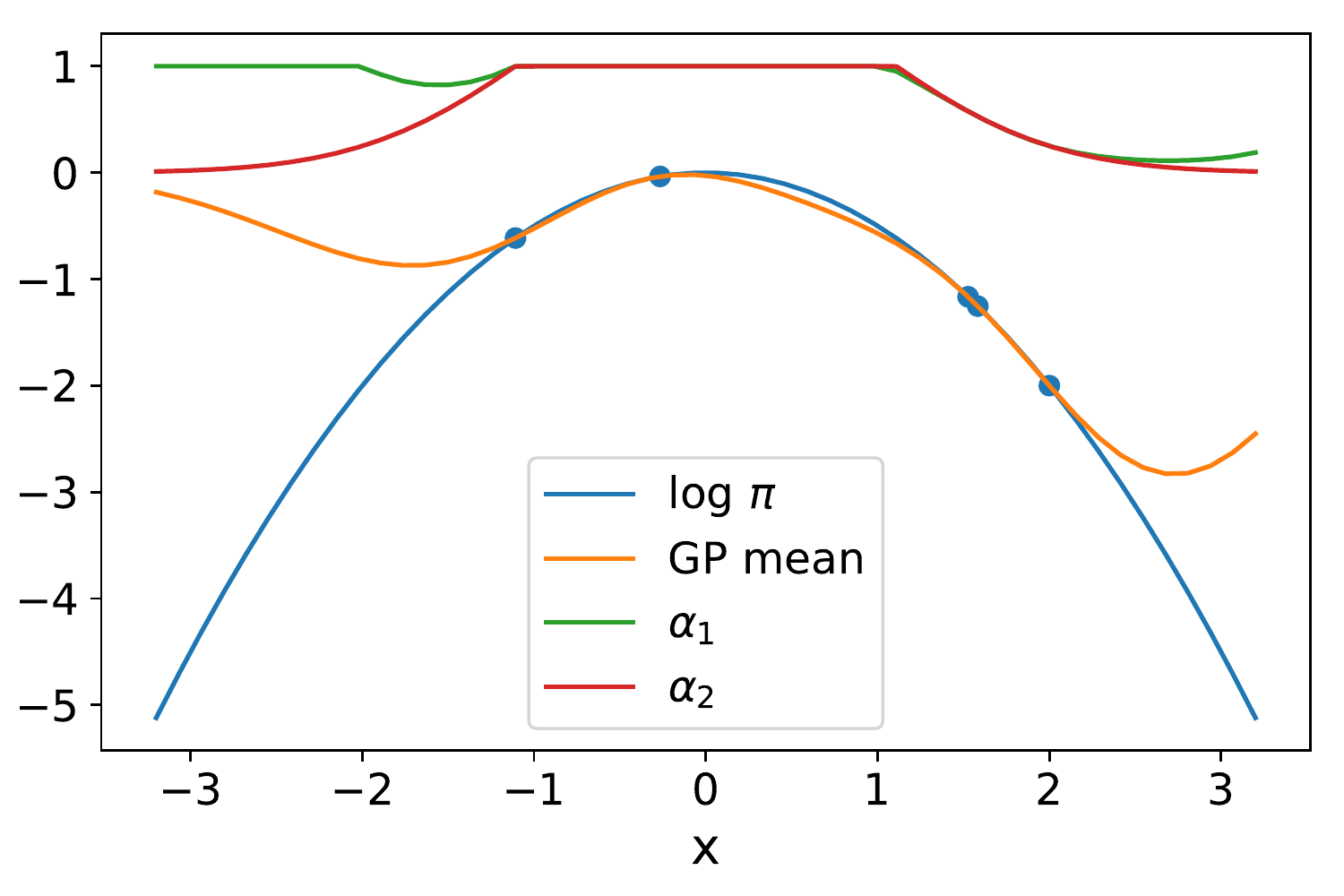}
  \caption{}
  \label{fig:sfig4}
\end{subfigure}
\begin{subfigure}{.32\textwidth}
  \centering
  \includegraphics[width=.8\linewidth]{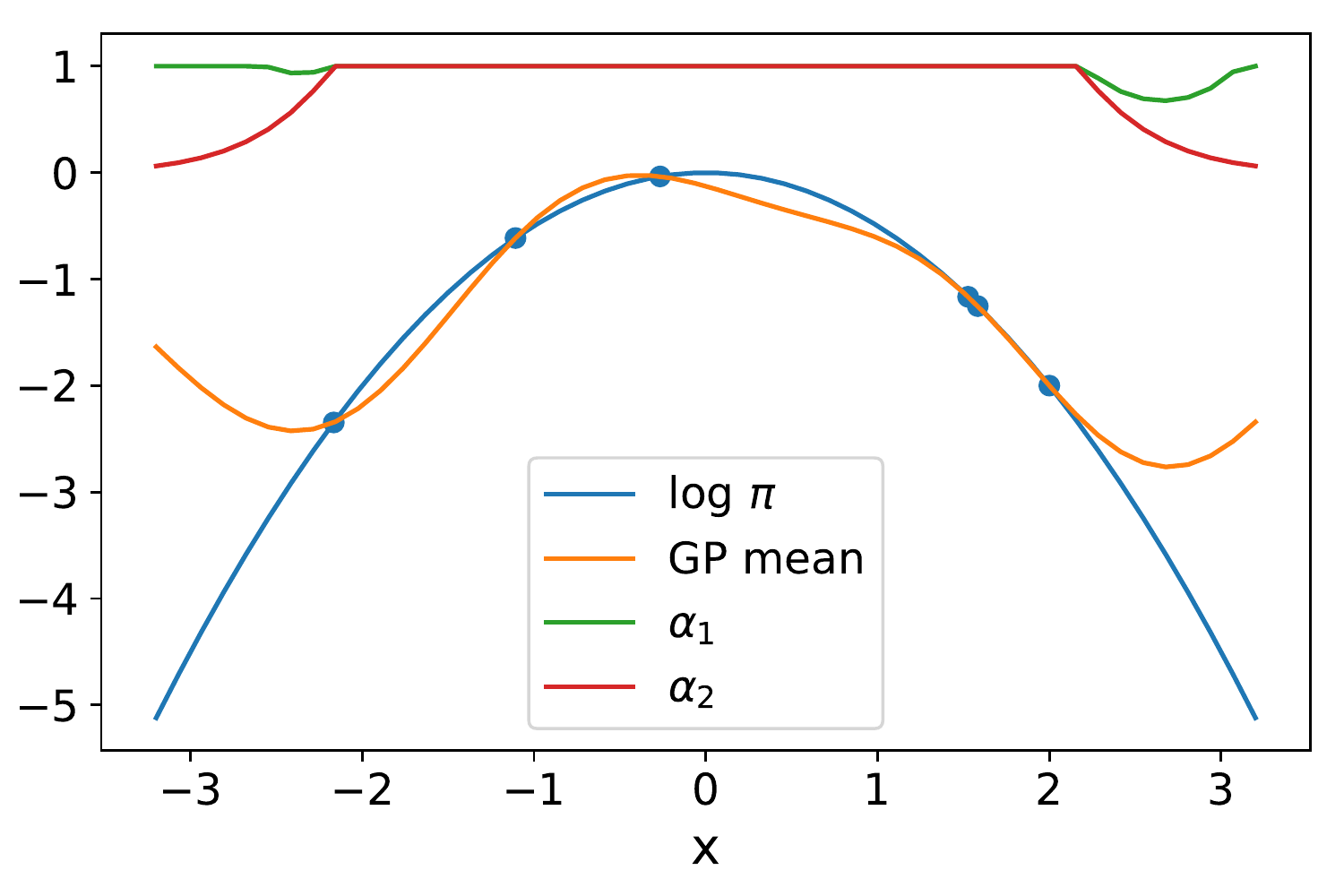}
  \caption{}
  \label{fig:sfig5}
\end{subfigure}
\begin{subfigure}{.32\textwidth}
  \centering
  \includegraphics[width=.8\linewidth]{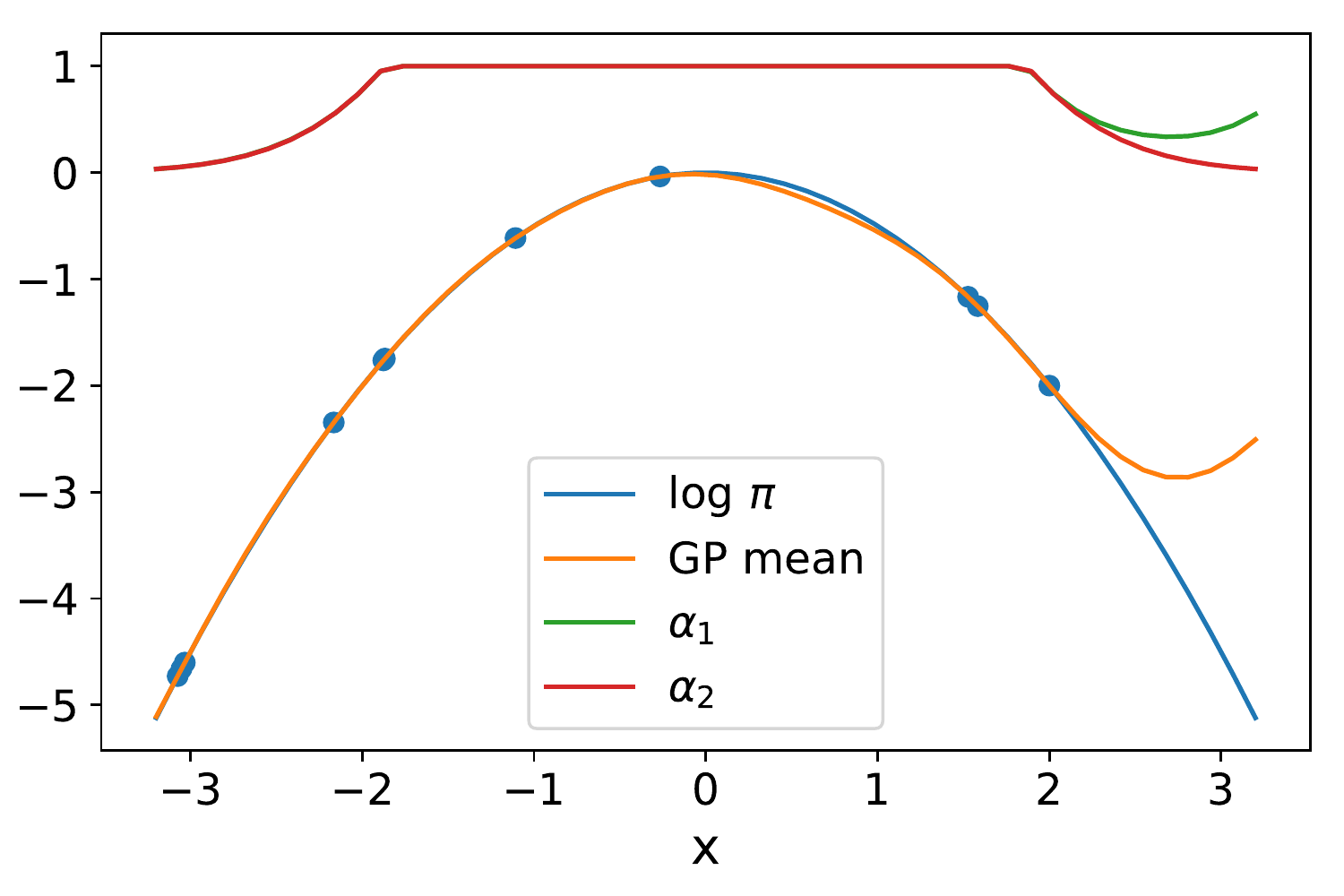}
  \caption{}
  \label{fig:sfig6}
\end{subfigure}
\caption{Convergence of the GP mean to the log-normal unnormalised density and of $\alpha_1$ to $\alpha_2$ with increasing iterations (a)-(f).}
\label{fig:1}
\end{figure}

For illustration, in Figure \ref{fig:1}, we consider a 1D case with $\pi(\theta)\propto e^{-\frac{x^2}{2}}$. It can be noticed how $\alpha_1$ converges to $\alpha_2$ at the increase of the log-likelihood evaluations in $\mathcal{I}_t$.

Proposition \ref{prop:5} holds under the assumption of fixed hyperparameters for the covariance function of the GP. Therefore, in our algorithm, we update the hyperparameters only during \textit{burnin}.

In the next section, we extend these results to Metropolis-adjusted Langevin method.

\section{Metropolis-adjusted Langevin} \label{sec:MALA}
The MALA takes one step (of step size $\delta>0$) in the direction of the gradient from the current point
\begin{equation}
\label{eq:MALAprop}
{\boldsymbol \theta}^*:={\boldsymbol \theta}^{(k)}+\frac{1}{2}\delta \Lambda \Big(\nabla LL({\boldsymbol \theta}^{(k)})+\nabla\log p({\boldsymbol \theta}^{(k)})\Big)+{\sqrt {\delta \Lambda}}\mathbf{z}
\end{equation}
with $\mathbf{z} \sim N(0,I)$ and $\Lambda$ is a preconditioning covariance matrix. Here $\sqrt{\Lambda}$ denotes the matrix square root. In this case, we assume that we can evaluate both the log-likelihood and its gradient
$\mathcal{I}_t=\{({\boldsymbol \theta}^{(i)},[\text{LL}(D|{\boldsymbol \theta}^{(i)}),\nabla^T\text{LL}(D|{\boldsymbol \theta}^{(i)})]): ~~i=1,\dots,t\}$. We use a multiple-output joint GP~\cite{Rasmussen02} as surrogate model for the log-likelihood and its gradient. The idea in this case is  simply to apply the previous two-stage algorithm 
using the proposal \eqref{eq:MALAprop} with gradient replaced by $\widetilde{\nabla LL}$. 

\begin{align}
\label{eq:alphastarMALA}
\tilde{\alpha}^{(1)}({\boldsymbol \theta}^{(k)},{\boldsymbol \theta}^*)= 1 \wedge \frac{\exp( \widetilde{\text{LL}}_t({\boldsymbol \theta}^*))p({\boldsymbol \theta}^*)q({\boldsymbol \theta}^{(k)}|{\boldsymbol \theta}^*+\frac{1}{2}\delta \Lambda \widetilde{\nabla LL}({\boldsymbol \theta}^*)+\frac{1}{2}\delta \Lambda\nabla\log p({\boldsymbol \theta}^*))}{\exp( \widetilde{\text{LL}}_t({\boldsymbol \theta}^{(k)}))p({\boldsymbol \theta}^{(k)})q({\boldsymbol \theta}^*|{\boldsymbol \theta}^{(k)}+\frac{1}{2}\delta \Lambda \widetilde{\nabla LL}({\boldsymbol \theta}^{(k)})+\frac{1}{2}\delta \Lambda\nabla\log p({\boldsymbol \theta}^{(k)}))}
\end{align}
where $q$ is the Normal proposal with covariance $\delta \Lambda$. Note that, $\widetilde{\text{LL}}_t({\boldsymbol \theta}^{(k)})),\widetilde{\nabla LL}({\boldsymbol \theta}^{(k)})$ are exact evaluations because of Assumption \ref{as1}. As before we can marginalise out $\widetilde{LL},\widetilde{\nabla LL}$ computing the expectation of $\alpha_1$ w.r.t.\ the GP. We use the following result.
\begin{proposition}
 \label{prop:5}
 The expectation of
 \begin{align}
 \label{eq:ppdf}
 e^{f({\boldsymbol \theta}^*)} q({\boldsymbol \theta}'|{\boldsymbol \theta}^*+\tfrac{\delta}{2}\Lambda\nabla f({\boldsymbol \theta}^*)+\tfrac{\delta}{2}\Lambda\nabla \log p({\boldsymbol \theta}^*) )
\end{align}
 w.r.t.\ the GP,  where $f,\nabla f$ denote the  GP distributed log-likelihood and its gradient and $p({\boldsymbol \theta}^*)$ is the prior, is equal to
  $$
  e^{V^T \begin{bmatrix}
    \mu\\ 
    \Lambda \mu_{\nabla}                  
\end{bmatrix}+\frac{1}{2}V^TKV +\frac{1}{2}\frac{\delta^2}{4} \mu_{\nabla}^T  \Lambda  \mu_{\nabla}} q({\boldsymbol \theta}'|{\boldsymbol \theta}^*+\tfrac{\delta}{2}\Lambda\nabla \log p({\boldsymbol \theta}^*) )
   $$ 
   with  $
 V=\left[1,\left({\boldsymbol \theta}'-{\boldsymbol \theta}^*-
 \tfrac{\delta}{2}\Lambda\nabla \log p({\boldsymbol \theta}^*)
 -\tfrac{\delta}{4}(\Lambda\mu_{\nabla})^T) \right)^T \Lambda^{-1}\tfrac{\delta}{2}\right]^T
 $, $\mu, \mu_{\nabla}$ are the GP predictive means for  $f,\nabla f$ and $K$ is the relative covariance matrix.
\end{proposition}
Stage 2 uses the exact evaluation of the log-likelihood and its gradient. We omit the details. The overall algorithm is similar to the one presented previously for MH with the only difference that the GP is multi-output over the log-likelihood and its gradient.

\section{Numerical experiments} \label{sec:examples}
To model the log-likelihood (and its gradient for MALA), we use a GP with Square Exponential covariance function. A zero mean is used  with the value of  $\text{LL}({\boldsymbol \theta}^{(k)})$ (and its gradient for MALA) subtracted. This is equivalent to defining a GP with prior mean equal to $\text{LL}({\boldsymbol \theta}^{(k)})$; in this way, far from the data, the acceptance probability $\alpha_1$ only depends on the variance of the GP. This guarantees a high probability of acceptance in Stage 1 for samples in large-uncertainty regions.  
The GP is initialised using  $3$ observations, before starting the two-stage sampler.

We consider five target distributions.
\begin{description}
 \item[T1] The 2D posterior  of the parameters $a,b$  of the banana shape distribution (true value set to $a=0.2,b=2$);
 \item[T2] The 3D posterior  of the parameters $a,b,\sigma$ of the nonlinear regression model $y=a\frac{x}{x+b}+\epsilon$, $\epsilon \sim N(0,\sigma^2)$ (true value set to $a=0.14,b=50,\sigma=0.1$).
   \item[T3] The 3D posterior  of the parameters 
   $\ell_1,\ell_2,\sigma^2$ of the SE kernel for a GP-classifier.
  \item[T4] The 4D posterior  of the parameters 
  $\beta,\gamma, \sigma_1,\sigma_2$ of a Susceptible, Infected, Recovery (SIR) model.
  \item[T5] The 5D posterior  of the parameters 
  $\beta_0,\dots,\beta_4$ of a parametric logistic regression problem.
\end{description}
Appendix \ref{app:priors} gives further details on priors assumed for the parameters and selected proposal. Each of these five posteriors has a specific feature, resulting in a diverse set of challenging targets, for instance T1  is heavy tailed and T2 is heavily anisotropic. T4, the SIR problem, is a prototypical example of the type of applications targeted by the proposed method. To compute the likelihood, we need to solve numerically a system of ODEs and, in more complex biological and chemical models, this can be computationally heavy. 

Evaluating the likelihood in these five problems is very fast, this allows us to quickly perform Monte Carlo simulations to assess the performance of the model by generating artificial data. We then evaluate the efficiency of the algorithms by simply counting the number of likelihood evaluations.

We compare our two-stage algorithm with the standard implementations of MH and MALA.
For each target problem and in each simulation, we generate 2500 samples (including 500 for burnin). We have deliberately selected a small number of samples to show that our approach converges quickly, which is important in computationally expensive applications.
We check for convergence to the correct posterior distribution using the metrics described in the caption of Table \ref{tab:res}.

Table \ref{tab:res} reports the value of the metrics averaged over the 30 simulations and over parameters.
Comparing the simulations' results it can be noticed that the proposed GP-based samplers obtain the same convergence metrics of the standard MH and MALA, but with a fraction of the number of likelihood evaluations. It can also be noticed how the fraction of the number of full likelihood evaluations required is problem dependent, ranging from 15\% for T4 to 65\% for T3. This demonstrates that our approach automatically adapts to the complexity of the specific target distribution.

\begin{table}
\begin{minipage}{0.45\textwidth}
{\scriptsize
 \begin{tabular}{ccccccc}
 & & AR& ESS & ESJD &  Eval\% & SD \\
 \hline
\multirow{4}{*}{T1} & MH & 0.37& 90 & 0.13 & 100 & 0.02 \\ 
& GP-MH & 0.36 & 113 & 0.13 & {\bf 41} & 0.02\\ 
& MALA& 0.26 & 73 & 0.2 & 100 & 0.03\\ 
& GP-MALA& 0.26 & 75 & 0.2 & {\bf 35} & 0.02\\
 \hline
\multirow{4}{*}{T3} & MH & 0.42& 137 & 0.44 & 100 & 4.1 \\ 
& GP-MH & 0.42 & 135 & 0.38 & {\bf 42} & 3.5\\ 
& MALA & 0.44& 133 & 0.48 & 100 & 4.1 \\ 
& GP-MALA & 0.43 & 134 & 0.45 & {\bf 45} & 3.5\\
 \hline
\multirow{4}{*}{T5} & MH & 0.29& 98 & 0.002 & 100 & 0.006 \\ 
& GP-MH & 0.29 & 102 & 0.002 & {\bf 35} & 0.006\\ 
& MALA & 0.67 & 339 & 0.009 & 100 & 0.006 \\  
& GP-MALA & 0.67 & 368 & 0.009 & {\bf 68} & 0.006\\ 
 \end{tabular}
}
\end{minipage}
\begin{minipage}{0.45\textwidth}
{\scriptsize
 \begin{tabular}{ccccccc}
 & & AR& ESS & ESJD &  Eval\% & SD \\
 \hline
\multirow{4}{*}{T2} & MH & 0.28& 138 & 32.6 & 100 & 339 \\ 
& GP-MH & 0.27 & 133 & 31 & {\bf 39} & 339\\ 
& MALA& 0.26 & 220 & 51 & 100 & 316\\ 
& GP-MALA& 0.21 & 147 & 29 & {\bf 43}  & 255\\
 \hline
\multirow{4}{*}{T4} & MH & 0.1& 51 & 0.003 & 100 & 0.009 \\ 
& GP-MH & 0.1 & 45 & 0.003 & {\bf 15} & 0.009\\ 
 &  & &  & &  &  \\ 
 &  & &  & &  &  \\ 
 \hline
\multirow{4}{*}{} &  & &  & &  &  \\ 
 &  & &  & &  &  \\ 
 &  & &  & &  &  \\ 
 &  & &  & &  &  \\ 
 \end{tabular}
}
\end{minipage}
\caption{Comparison between the proposed GP-based samplers and standard MH and MALA in terms of the acceptance rate (AR), Effective Sample Size (ESS),  
Expected Square Jumping Distance (ESJD), percentage of likelihood evaluations in the 2000 iterations (EVAL), Square Distance (SD) between the true value of the parameters and the estimated posterior mean. We have not run MALA for the SIR model.}
\label{tab:res}
\end{table}

\section{Conclusions}
We have presented a two-stage Metropolis-Hastings algorithm for sampling probabilistic models, whose log-likelihood is computationally expensive to evaluate, by using a surrogate GP model.
The key feature of the approach, and the difference w.r.t.\ previous works, is the ability to learn the target distribution  from scratch (while sampling), and so without the need of pre-training the GP. This is  fundamental for automatic and  inference in Probabilistic Programming Languages
In particular, we have presented an alternative first stage acceptance scheme by marginalising out the GP distributed function, which makes the acceptance ratio explicitly dependent  on the variance of the GP. This approach is extended to Metropolis-Adjusted Langevin algorithm (MALA). Numerical experiments have demonstrated the effectiveness of the method, which can automatically adapt to the complexity of the target distribution. In the numerical experiments, we have used a full GP whose computational load grows cubically as the size of the training set increases. Sparse GPs  can be employed to address this issue \citep{quinonero2005unifying,snelson2006sparse,pmlrv5titsias09a,Hensman2013,hernandez2016scalable,bauer2016understanding,SCHURCH2020} when it is necessary to sample thousands of samples.

In future work, we plan to extend the approach we used for MALA to Hamiltonian Monte Carlo. We also intend to investigate whether tailored covariance functions for log densities or ratios of densities can provide any convergence advantage, but also investigate surrogate models alternative to GPs.

\begin{acks}
\end{acks}

\bibliographystyle{acm-reference-format}
\bibliography{biblio}

\newpage
\appendix

\section{Priors for T1--T5}
\label{app:priors}
A zero prior is used for T1. In each simulation, we generate a different starting point in the interval $[-2,2]$.

For T2, the prior is $a \sim N(3,1), b \sim N(30,15^2), \log \sigma \sim N(-2,1)$ and $x=[28, 55,83, 110, 138, 225, 375]$. In each simulation, we generate different $y$ according to the likelihood reported in the main text.
 
For T3, the prior is $\log {\ell_i}, \log \sigma \sim N(0,10)$. The GP classification dataset includes 1000 points with $x_1,x_2 \sim N(0,1)$. In each simulation,
the true two lengthscales are uniformly sampled from $[0.1,1]$. The true variance is fixed at $\sigma^2=5$. 

For T4, we use the probabilistic model
described in \cite{PyMC3ODE}. 

For T5, the prior is $\beta_i \sim N(0,100)$. The GP classification dataset includes 1000 points with $x_1,x_2 \sim N(0,1)$. In each simulation,
the true $\beta_i$ are sampled from  $N(0,1)$.

In all cases, we have used Normal proposals for MH with diagonal covariance matrix.

\section{Proofs}
\label{app:proofs}

\subsection{Proof of Proposition \ref{prop:1}}
This follows by the mean of the log-normal distribution.

\subsection{Proof of Proposition \ref{prop:2}}
We prove detailed balance:
\begin{align*}
&e^{\mu({\boldsymbol \theta}^{(k)}|\mathcal{I}_t) +\frac{1}{2}k({\boldsymbol \theta}^{(k)},{\boldsymbol \theta}^{(k)}|\mathcal{I}_t)}p({\boldsymbol \theta}^{(k)})q({\boldsymbol \theta}^*|{\boldsymbol \theta}^{(k)} ) \alpha^{(1)}({\boldsymbol \theta}^{(k)},{\boldsymbol \theta}^*)\\
&=e^{\mu({\boldsymbol \theta}^{(k)}|\mathcal{I}_t) +\frac{1}{2}k({\boldsymbol \theta}^{(k)},{\boldsymbol \theta}^{(k)}|\mathcal{I}_t)}p({\boldsymbol \theta}^{(k)})q({\boldsymbol \theta}^*|{\boldsymbol \theta}^{(k)} ) 
\left( 1 \wedge \tfrac{e^{\mu({\boldsymbol \theta}^*|\mathcal{I}_t) +\frac{1}{2}k({\boldsymbol \theta}^*,{\boldsymbol \theta}^*|\mathcal{I}_t)}p({\boldsymbol \theta}^*)q({\boldsymbol \theta}^{(k)} |{\boldsymbol \theta}^*)}{e^{\mu({\boldsymbol \theta}^{(k)}|\mathcal{I}_t) +\frac{1}{2}k({\boldsymbol \theta}^{(k)},{\boldsymbol \theta}^{(k)}|\mathcal{I}_t)}p({\boldsymbol \theta})q({\boldsymbol \theta}^*|{\boldsymbol \theta}^{(k)})}\right)\\
&= \left( e^{\mu({\boldsymbol \theta}^{(k)}|\mathcal{I}_t) +\frac{1}{2}k({\boldsymbol \theta}^{(k)},{\boldsymbol \theta}^{(k)}|\mathcal{I}_t)}p({\boldsymbol \theta}^{(k)})q({\boldsymbol \theta}^*|{\boldsymbol \theta}^{(k)} ) \wedge e^{\mu({\boldsymbol \theta}^*|\mathcal{I}_t) +\frac{1}{2}k({\boldsymbol \theta}^*,{\boldsymbol \theta}^*|\mathcal{I}_t)}p({\boldsymbol \theta}^*)q({\boldsymbol \theta}^{(k)} |{\boldsymbol \theta}^*)\right)\\
 &= \left( \tfrac{e^{\mu({\boldsymbol \theta}^{(k)}|\mathcal{I}_t) +\frac{1}{2}k({\boldsymbol \theta}^{(k)},{\boldsymbol \theta}^{(k)}|\mathcal{I}_t)}p({\boldsymbol \theta}^{(k)})q({\boldsymbol \theta}^*|{\boldsymbol \theta}^{(k)} )}{e^{\mu({\boldsymbol \theta}^*|\mathcal{I}_t) +\frac{1}{2}k({\boldsymbol \theta}^*,{\boldsymbol \theta}^*|\mathcal{I}_t)}p({\boldsymbol \theta}^*)q({\boldsymbol \theta}^{(k)} |{\boldsymbol \theta}^*)} \wedge 1\right)e^{\mu({\boldsymbol \theta}^*|\mathcal{I}_t) +\frac{1}{2}k({\boldsymbol \theta}^*,{\boldsymbol \theta}^*|\mathcal{I}_t)}p({\boldsymbol \theta}^*)q({\boldsymbol \theta}^{(k)} |{\boldsymbol \theta}^*).
\end{align*}
which ends the proof. The term in brackets in the last equation is  $\alpha^{(1)}({\boldsymbol \theta}^*,{\boldsymbol \theta}^{(k)})$. Because of Assumption \ref{as1}, we have that
$k({\boldsymbol \theta}^{(k)},{\boldsymbol \theta}^*|\mathcal{I}_t)=0$ (independent). This is the reason we can assume work on the numerator and denominator independently. 

\subsection{Proof of Proposition \ref{prop:3}}
Let ${\boldsymbol \theta}'$ denote the new point in $\mathcal{I}_{t-1}$ and with $R$ all the points in $\mathcal{I}_{t-1}$, by definition of Kernel matrix:
 \begin{align}
K_t = \begin{bmatrix}
K_{t-1}  & k_{t-1}({\boldsymbol \theta}')\\k_{t-1}({\boldsymbol \theta}')^\top & k({\boldsymbol \theta}',{\boldsymbol \theta}')
\end{bmatrix}:=
\begin{bmatrix}
k(R,R)  & k({\boldsymbol \theta}',R)\\k({\boldsymbol \theta}',R)^\top & k({\boldsymbol \theta}',{\boldsymbol \theta}')
\end{bmatrix}.
\end{align}
The predicted variance at step $t$ at the point ${\boldsymbol \theta}^*$ is
$$
k({\boldsymbol \theta}^*, {\boldsymbol \theta}^*)-k_t({\boldsymbol \theta}^*)^\top K_t^{-1}k_t({\boldsymbol \theta}^*),
$$
while the predicted variance at step $t-1$ at the point ${\boldsymbol \theta}^*$ is
$$
k({\boldsymbol \theta}^*, {\boldsymbol \theta}^*)-k_{t-1}({\boldsymbol \theta}^*)^\top K_{t-1}^{-1}k_{t-1}({\boldsymbol \theta}^*)
$$
Therefore, we need to prove that
$$
k_t({\boldsymbol \theta}^*)^\top K_t^{-1}k_t({\boldsymbol \theta}^*)>k_{t-1}({\boldsymbol \theta}^*)^\top K_{t-1}^{-1}k_{t-1}({\boldsymbol \theta}^*).
$$
The inverse of $K_t$ is:
\begin{align}
K_t^{-1} = \begin{bmatrix}K_{t-1}^{-1}  + K_{t-1}^{-1}  k_{t-1}({\boldsymbol \theta}')Mk_{t-1}({\boldsymbol \theta}')^\top K_{t-1}^{-1} & -K_{t-1}^{-1} k_{n-1}({\boldsymbol \theta})M \\ -Mk_{t-1}({\boldsymbol \theta})^\top K_{t-1}^{-1} & M \end{bmatrix},
\end{align}
with 
$M  = (k({\boldsymbol \theta}', {\boldsymbol \theta}') - k_{t-1}({\boldsymbol \theta}')^\top K_{t-1}^{-1}k_{t-1}({\boldsymbol \theta}'))^{-1}$.
Now note that
\begin{align*}
 &k_t({\boldsymbol \theta}^*)^\top K_t^{-1}k_t({\boldsymbol \theta}^*)\\
 &=\begin{bmatrix}
  k^\top_{t-1}({\boldsymbol \theta}^*), k'({\boldsymbol \theta}^*)                                                                                                                                           \end{bmatrix}
  \begin{bmatrix}K_{t-1}^{-1}  + K_{t-1}^{-1}  k_{t-1}({\boldsymbol \theta}')Mk_{t-1}({\boldsymbol \theta}')^\top K_{t-1}^{-1} & -K_{t-1}^{-1} k_{n-1}({\boldsymbol \theta})M \\ -Mk_{t-1}({\boldsymbol \theta})^\top K_{t-1}^{-1} & M \end{bmatrix}
  \begin{bmatrix}
  k_{t-1}({\boldsymbol \theta}^*)\\
  k'({\boldsymbol \theta}^*)                                                                                                                                           \end{bmatrix}
\end{align*}
which is equal to:
\begin{align*}
k_t({\boldsymbol \theta}^*)^\top K_t^{-1}k_t({\boldsymbol \theta}^*) & = k_{t-1}^\top({\boldsymbol \theta}^*) K_{t-1}^{-1} k_{t-1}({\boldsymbol \theta}^*) + k_{t-1}^\top({\boldsymbol \theta}^*) K_{t-1}^{-1} k_{t-1}({\boldsymbol \theta}')Mk_{t-1}({\boldsymbol \theta}')^\top K_{t-1}^{-1} k_{t-1}({\boldsymbol \theta}^*)\\
& - k'({\boldsymbol \theta}^*)Mk_{t-1}({\boldsymbol \theta}')^\top K_{t-1}^{-1} k_{t-1}({\boldsymbol \theta}^*) - k_{t-1}({\boldsymbol \theta}^*)^\top K_{t-1}^{-1} k_{t-1}({\boldsymbol \theta}')Mk'({\boldsymbol \theta}^*) + k'({\boldsymbol \theta}^*)Mk'({\boldsymbol \theta}^*).
\end{align*}
where $k'({\boldsymbol \theta}^*):=k({\boldsymbol \theta}^*,{\boldsymbol \theta}')$. Therefore, we have that
\begin{align*}
k_t({\boldsymbol \theta}^*)^\top K_t^{-1}k_t({\boldsymbol \theta}^*) & = k_{t-1}^\top({\boldsymbol \theta}^*) K_{t-1}^{-1} k_{t-1}({\boldsymbol \theta}^*) + N.
\end{align*}
with 
$$
N=M(k_{t-1}^\top({\boldsymbol \theta}^*) K_{t-1}^{-1} k_{t-1}({\boldsymbol \theta}') - k'({\boldsymbol \theta}^*))^2=M(k_{t-1}^\top({\boldsymbol \theta}^*) K_{t-1}^{-1} k_{t-1}({\boldsymbol \theta}') - k({\boldsymbol \theta}^*,{\boldsymbol \theta}'))^2
$$
which is strictly greater than zero whenever 
${\boldsymbol \theta}^*,{\boldsymbol \theta}' \notin \mathcal{I}_{t-1}$.  Note in fact that, under Assumption \ref{as1}, $N=0$ only if either ${\boldsymbol \theta}^* \in \mathcal{I}_{t-1}$ or ${\boldsymbol \theta}' \in \mathcal{I}_{t-1}$ (exact interpolation property). If the proposal distribution is  absolutely continuous
w.r.t.\ the Lebesgue measure on $\Theta$, then   ${\boldsymbol \theta}^*,{\boldsymbol \theta}' \notin \mathcal{I}_{t-1}$  holds with probability 1.

\subsection{Proof of Proposition \ref{prop:5}}
 We work in log-scale and omit the dependence
 of $f,\nabla f$ on ${\boldsymbol \theta}^*$ for notation simplification and so we can rewrite the product of the two PDF in \eqref{eq:ppdf}:
 \begin{align*}
&  f-\frac{1}{2}\left({\boldsymbol \theta}'-{\boldsymbol \theta}^*-
 \frac{\delta}{2}\Lambda\nabla \log p({\boldsymbol \theta}^*)
 \right)^T \Lambda^{-1}(\cdots)+\frac{2}{2}\left({\boldsymbol \theta}'-{\boldsymbol \theta}^*-
 \frac{\delta}{2}\Lambda\nabla \log p({\boldsymbol \theta}^*)
 \right)^T \Lambda^{-1}(
   \frac{\delta}{2}\Lambda\nabla f)\\
 &-\frac{1}{2}\left(
   \frac{\delta}{2}\Lambda \nabla f
 \right)^T \Lambda^{-1}(\cdots)-\frac{1}{2}\begin{bmatrix}
    f-\mu, \nabla^T f-\mu^T_{\nabla}                        
\end{bmatrix}\begin{bmatrix}
A & B\\
B^T & C
\end{bmatrix}
 \begin{bmatrix}
    f-\mu\\ 
    \nabla f-\mu_{\nabla}                     
\end{bmatrix}
 \end{align*}
  where $\Lambda$ is the covariance matrix of the proposal, and the last term in the above sum is the GP predictive posterior with mean $[\mu,\mu_{\nabla}  ]$ and covariance $K$. We have expressed $K^{-1}$ as the block matrix with blacks $A,B,C,D$.
We can rewrite the above sum as
 \begin{align*}
&  f-\frac{1}{2}\left({\boldsymbol \theta}'-{\boldsymbol \theta}^*-
 \frac{\delta}{2}\Lambda\nabla \log p({\boldsymbol \theta}^*)
 \right)^T \Lambda^{-1}(\cdots)+\frac{2}{2}\left({\boldsymbol \theta}'-{\boldsymbol \theta}^*-
 \frac{\delta}{2} \Lambda\nabla \log p({\boldsymbol \theta}^*)
 \right)^T \Lambda^{-1}(
   \frac{\delta}{2}\Lambda \nabla f)\\
 &-\frac{2}{2}\frac{\delta^2}{4} \mu_{\nabla}^T  \Lambda \nabla f+\frac{1}{2}\frac{\delta^2}{4} \mu_{\nabla}^T  \Lambda  \mu_{\nabla}-\frac{1}{2}\begin{bmatrix}
    f-\mu, \nabla^T f - \mu^T_{\nabla}                      
\end{bmatrix}\begin{bmatrix}
A & B\\
B^T & \frac{\delta^2}{4}\Lambda+C
\end{bmatrix}
 \begin{bmatrix}
    f-\mu\\ 
    \nabla f -\mu_{\nabla}                  
\end{bmatrix}
 \end{align*}
 We can now define the vector $V$  and prove the result using
 the moments of the multivariate Lognormal distribution \citep{halliwell2015lognormal}.

\end{document}